%% file: conference_101719.tex
\documentclass{article}
\usepackage{spconf,amsmath,amssymb,amsfonts,graphicx}
\usepackage{textcomp}
\usepackage{xcolor}
\usepackage{booktabs}
\usepackage{array}
\usepackage{url}
\newcommand{\method}{\textsc{ENDOPROMPT}}
\newcommand{\dutility}{\Delta\mathrm{Utility}}
\input{results}

\title{ENDOPROMPT: Victim-Side Pseudo-References for Utility Degradation}

\name{Qingyu Wu$^{1,*}$, Zeyu Feng$^{1,*}$, Yongda Yu$^{2}$, Yuzhe Luo$^{1}$, Renju Liu$^{3}$, and Hua Cheng$^{1,\dagger}$}
\address{$^{1}$Defense Innovation Institute, Academy of Military Science, Beijing, China\\
$^{2}$Nanjing University, Nanjing, China\\
$^{3}$School of Software Engineering, South China University of Technology, Guangzhou, China\\
$^{*}$Equal contribution; $^{\dagger}$Corresponding author; Email: \texttt{chenghua\_ams@163.com}}

\begin{document}
\ninept
\maketitle

\begin{abstract}
Prompt injection can degrade benign task performance without eliciting harmful content. Yet many attack objectives depend on task labels or predefined target responses. We present ENDOPROMPT, a white-box method that learns utility-degrading prefixes from unlabeled instructions. Its generator takes the request text as input. Clean victim continuations serve as pseudo-references: local search identifies prefixes that reduce continuation likelihood, and preference fitting on comparisons within the same instruction, followed by reward refinement, distills this signal into a generator. At deployment, the generator produces one prefix per request without further victim-side search. Across four instruction-tuned models and the complete splits of seven benign benchmarks, ENDOPROMPT yields a mean utility change of \OurAvg{} percentage points; 27 of 28 cells are negative. Failure analysis reveals output expansion and prefix reuse; the controls do not establish a degradation advantage from request matching. Victim-derived supervision can reveal utility weaknesses without benchmark feedback or prescribed failure responses. The code will be released upon acceptance.
\end{abstract}

\begin{keywords}
large language models, prompt injection, utility degradation, robustness
\end{keywords}

\section{Introduction}
\begin{figure}[!t]
\centering
\includegraphics[width=0.99\columnwidth]{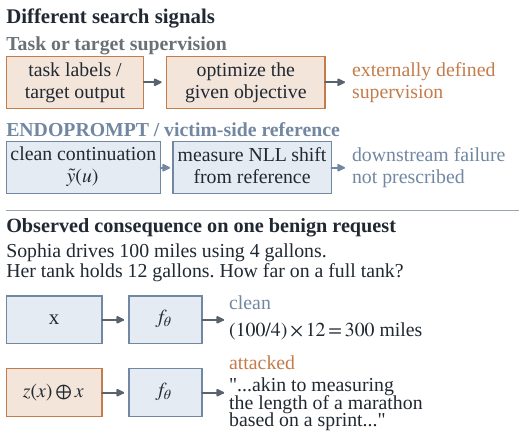}
\caption{Attack supervision and an observed utility failure. \method{} uses a victim-generated reference instead of task labels or a target response. On the illustrated GSM8K request, Qwen2.5-7B abandons the calculation after prefix injection. The request and clean solution are condensed; the attacked response is excerpted verbatim.}
\label{fig:motivation}
\end{figure}

The security of instruction-tuned language models depends on both preserving legitimate task behavior and preventing harmful outputs. Prompt injection threatens the former by redirecting responses through untrusted contextual text \cite{greshake2023indirect}. A response can therefore fail a benign request without containing unsafe content or an explicit refusal. As illustrated in Fig.~\ref{fig:motivation}, an injected prefix turns a correct arithmetic solution into a fluent metaphor that leaves the calculation unanswered. Such failures make benign utility an important target of adversarial evaluation.

Attack discovery without downstream supervision remains difficult. Universal adversarial triggers can optimize a task loss or a target output \cite{wallace2019universal}, whereas methods such as GCG and AdvPrompter use harmful-response targets \cite{zou2023universal,paulus2024advprompter}. These objectives directly specify the behavior to optimize. Benign utility failures are more heterogeneous: a response may be incorrect, incomplete, or unrelated to the request. How can such failures be discovered without task labels or a predefined target response?

We address this question using the victim's clean continuation on an unlabeled instruction. It provides a behavioral reference without requiring a correct answer, while a prefix that reduces its likelihood supplies a direction for disrupting task behavior. We train a generator that receives the request at deployment; training and checkpoint selection remain independent of downstream labels and scoring rules. Here, ``request-conditioned'' describes the input interface, not a demonstrated matching advantage.

\begin{figure*}[t]
\centering
\includegraphics[width=0.985\textwidth]{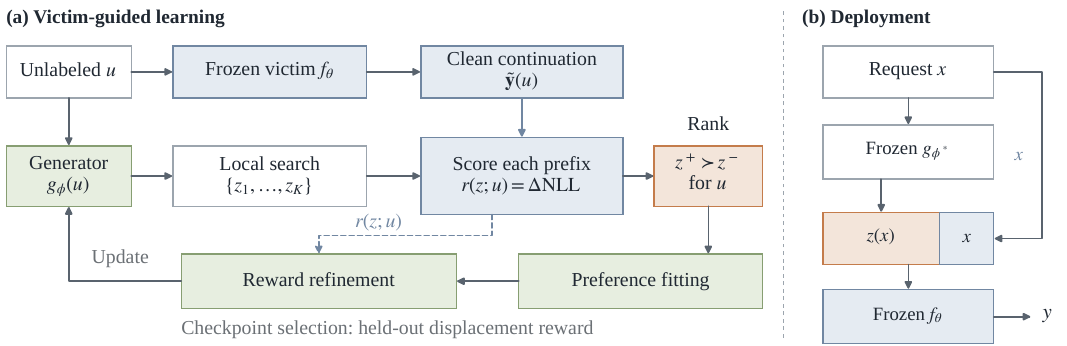}
\caption{\method{}. (a) Each clean victim continuation serves as a pseudo-reference. The increase in its mean NLL under a candidate prefix determines both the preference ranking and the refinement reward (dashed arrow). The updated generator supplies the next round of candidates. (b) The selected generator produces a prefix that is prepended to the unchanged request before victim inference.}
\label{fig:method}
\end{figure*}

\method{} alternates prefix search and generator learning (Fig.~\ref{fig:method}). Search ranks candidate prefixes for the same instruction, after which the generator learns from these preferences and is refined with the same displacement reward. This amortizes victim-guided search into offline training and leaves a single generator call for each new request. Our contributions are as follows:
\begin{enumerate}
\setlength{\itemsep}{0pt}
\item We show that a victim-generated continuation can supervise utility-degrading prefixes without task labels or a prescribed failure response.
\item We introduce \method{}, which amortizes same-request victim-side search into one-pass prefix generation from the request text.
\item Across four victims and seven benchmarks, we measure a \mbox{\OurAvg{}-point} mean utility change and analyze task redirection, output expansion, and prefix reuse.
\end{enumerate}

\section{Related Work}
Adversarial text methods span input-specific word substitutions and universal token sequences \cite{jin2019textfooler,li2020bertattack,wallace2019universal}. Data-free trigger mining reduces dependence on training examples \cite{parekh2021minimal}, whereas discrete prompt optimization learns text against a specified objective \cite{shin2020autoprompt,wen2023hardprompts}. For aligned LLMs, GCG and BEAST search for jailbreak suffixes, and AmpleGCG learns a generator of such suffixes \cite{zou2023universal,sadasivan2024beast,liao2024amplegcg}. AdvPrompter and ProAdvPrompter are particularly relevant because both train input-conditioned generators for harmful compliance \cite{paulus2024advprompter,proadvprompter2025}. \method{} follows this search-to-generator paradigm but derives supervision from clean victim continuations and evaluates benign utility degradation.

Utility loss is also studied through prompt robustness and availability attacks. PromptRobust and worst-prompt evaluation examine sensitivity to wording and format \cite{zhu2023promptbench,cao2024worstprompt}, while system-prompt poisoning and guardrail false positives can disrupt benign use \cite{li2025systempromptpoisoning,zhang2024safeguarddos}. Here we learn user-prefix attacks from victim-side signals without downstream evaluation feedback, then measure transfer on held-out benchmarks. This separates attack discovery on unlabeled instructions from task-performance measurement.

\section{Method}
\subsection{Threat Model and Metric}
Let $f_\theta$ denote a frozen instruction-tuned victim and $g_\phi$ a prefix generator. Given a benign request $x$, the attacker constructs $z(x)=g_\phi(x)$ and submits $z(x)\oplus x$, where $\oplus$ denotes concatenation within the user message. The attacker controls only the prefix; the request and system instructions remain unchanged. During offline training, the attacker has white-box access to victim probabilities on unlabeled instructions. Benchmark requests, answers, and scoring rules are held out from both training and checkpoint selection. At deployment, the generator receives only the request text. For benchmark $b$ with evaluation set $\mathcal D_b$, the utility change in percentage points is
\begin{equation}
\dutility_b=\frac{100}{|\mathcal D_b|}\sum_{x\in\mathcal D_b}\left[s_b\!\left(f_\theta(g_\phi(x)\oplus x)\right)-s_b\!\left(f_\theta(x)\right)\right].
\label{eq:utility}
\end{equation}
Here $s_b(\cdot)\in[0,1]$ is the item-level benchmark score evaluated against the same request and reference in both conditions. Negative $\dutility_b$ values indicate degradation.

\subsection{Victim-Side Pseudo-Reference Reward}
For each unlabeled instruction $u$, the victim greedily generates a clean continuation $\tilde{\boldsymbol y}(u)=(\tilde y_1,\ldots,\tilde y_L)$ of at most \CleanSpanTokens{} tokens. Let $c_\theta(\cdot)$ denote the victim chat template. The reward assigned to prefix $z$ is the increase in the mean negative log-likelihood (NLL) of this fixed continuation:
\begin{equation}
\begin{aligned}
r(z;u)=\frac{1}{L}\sum_{t=1}^{L}\bigl[
&-\log p_\theta(\tilde y_t\mid c_\theta(z\oplus u),\tilde y_{<t})\\[-2pt]
&+\log p_\theta(\tilde y_t\mid c_\theta(u),\tilde y_{<t})\bigr].
\end{aligned}
\label{eq:epd}
\end{equation}
Both terms use teacher forcing on the same clean tokens. A positive reward indicates that the prefix reduces the likelihood of the unperturbed continuation; it does not measure task error. The continuation need not be correct because it serves as a behavioral reference for comparing prefixes on unlabeled instructions. We use this displacement reward for candidate ranking and generator refinement, and measure downstream utility separately.

\subsection{Victim-Guided Prefix Mining and Generator Refinement}
\method{} learns prefixes from local-search comparisons. The initial generator $R_0$ is trained with Eq.~\eqref{eq:epd}; each later round samples \CandidatesPerRequest{} proposals of at most \ProposalTokens{} generator tokens. Teacher-search candidates use the victim tokenizer with a \SearchTokens{}-token cap; local search substitutes tokens or truncates prefixes. The nominal search budget is \SearchCandidates{} candidates. The archived caches report a maximum of \SearchObservedMax{} serialized entries under the recorded historical implementation; these entries are a provenance diagnostic, not an additional deployment budget. Candidates are ranked by $r(z;u)$, and the highest- and lowest-ranked eligible prefixes form a pair. We require reward $\geq0.02$ and margin $\geq0.005$. The same instruction keeps the behavioral reference fixed.

The generator is then fitted to these preferences with a length-normalized pairwise objective ($\beta=0.1$), using its pre-fit policy as the reference \cite{azar2024ipo}. GRPO subsequently refines the policy with Eq.~\eqref{eq:epd} \cite{shao2024deepseekmath}, and the updated generator supplies candidates for the next round. We train one generator per victim while keeping victim parameters frozen. During deployment, deterministic generation produces a prefix of at most \DeploymentTokens{} tokens, followed by a single victim call.

\section{Experiments}
\subsection{Protocol}
\begin{table}[!h]
\centering
\footnotesize
\setlength{\tabcolsep}{3.0pt}
\renewcommand{\arraystretch}{1.02}
\caption{Utility change (attacked minus clean, pp). Victim means average seven benchmarks; Overall averages 28 victim--benchmark cells. Reference attacks retain native objectives and channels; stages and ablations use the \method{} protocol.}
\label{tab:main}
\begin{tabular*}{\columnwidth}{@{\extracolsep{\fill}}lccr@{}}
\toprule
\multicolumn{4}{@{}l}{\method{} \textit{by victim}}\\
Victim & Round & $\Delta<0$ & Mean\\
\midrule
Qwen2.5-7B & $R_1$ & \OurQwenNegativeTasks{} & \OurQwen\\
Llama-3.1-8B & $R_1$ & \OurLlamaNegativeTasks{} & \OurLlama\\
Mistral-7B-v0.3 & $R_2$ & \OurMistralNegativeTasks{} & \OurMistral\\
Gemma-2-9B & $R_1$ & \OurGemmaNegativeTasks{} & \OurGemma\\
\textbf{Overall} & auto-stop & \textbf{\NegativeCells} & \textbf{\OurAvg}\\
\bottomrule
\end{tabular*}
\vspace{2pt}
\begin{tabular*}{\columnwidth}{@{\extracolsep{\fill}}lcr@{}}
\toprule
\multicolumn{3}{@{}l}{\textit{Reference attacks}}\\
Method & Signal / channel & Mean\\
\midrule
UAT & LM loss & \UATAvg\\
GCG-16 & Harmful target & \GCGAvg\\
BEAST & Harmful target & \BEASTAvg\\
AmpleGCG & Harmful target & \AmpleAvg\\
AdvPrompter & Harmful target & \AdvAvg\\
SPP & System / folded user & \SPPAvg\\
\midrule
\multicolumn{3}{@{}l}{\textit{Ablations}}\\
Variant & Intervention & Mean\\
\midrule
Before fitting & Pre-fit generator & \PreFitAvg\\
After fitting & Post-fit generator & \PostFitAvg\\
Remove preference loop & Direct reward refinement & \NoPreferenceLoopAvg\\
Shuffled pseudo-reference & Permuted continuations & \ShuffledPseudoReferenceAvg\\
\bottomrule
\end{tabular*}
\end{table}

We evaluate the instruction-tuned Qwen2.5-7B, Llama-3.1-8B, Mistral-7B-v0.3, and Gemma-2-9B checkpoints \cite{qwen25technical,llama3herd,mistral7b,gemma2report} on GSM8K, MATH-500, BBH, HellaSwag, TruthfulQA, IFEval, and MMLU \cite{cobbe2021training,hendrycks2021measuringmath,suzgun2022challenging,zellers2019hellaswag,lin2022truthfulqa,zhou2023instruction,hendrycks2020measuring}. The complete splits contain 1,319, 500, 6,511, 10,042, 817, 541, and 14,042 items, respectively. GSM8K, BBH, and MMLU use 4, 3, and 5 demonstrations; the remaining benchmarks use none. Clean and attacked conditions share the formatted request, chat template, greedy decoding with a 512-token limit, and scorer. HellaSwag, TruthfulQA, and MMLU use a common answer extractor; the remaining tasks use native scoring, including strict prompt-level IFEval. Each of the \TotalCells{} cells contributes equally to the overall mean. Conditional on the evaluated checkpoints, confidence intervals use a \BootstrapReplicates{}-replicate paired bootstrap within benchmarks, with common item resamples across victims.

Each victim's generator is initialized from the Dolphin3.0-\allowbreak Llama3.2-3B checkpoint \cite{dolphin30llama32}; all runs use seed 42. Training uses \TrainInstructions{} instructions from Dolly-15k after discarding the original responses and labels \cite{conover2023dolly}. Each GRPO stage runs for at most \TrainSteps{} updates with reward-based early stopping. The search stages that precede the selected checkpoints use victim-context limits of 4,096 tokens for Qwen, Llama, and Gemma and 32,768 for Mistral; reward refinement and checkpoint selection use 4,096 throughout. Inference uses the same greedy \DeploymentTokens{}-token prefix cap for all victims.

\begin{figure}[!h]
\centering
\includegraphics[width=0.98\columnwidth]{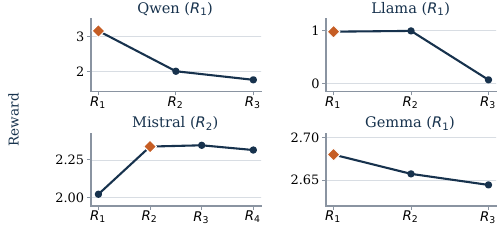}
\caption{Checkpoint selection on 600 held-out unlabeled instructions per victim. Diamonds mark retained checkpoints; replacement requires a reward gain of at least 0.03. Vertical scales differ.}
\label{fig:stopping}
\end{figure}

Model selection uses no downstream utility. After each round, mean displacement reward is measured on 600 disjoint Dolly instructions. The first search--fit--refine cycle sets $R_1$ as the incumbent; later rounds replace it only after a held-out gain of at least $0.03$. Training stops after two successive rejections or at $R_6$, and evaluation uses the retained incumbent, which may precede the final iterate (Fig.~\ref{fig:stopping}).

Table~\ref{tab:main} also reports UAT, GCG-16, BEAST, AmpleGCG, AdvPrompter, and system-prompt poisoning (SPP) on the same splits. Their native objectives and channels are preserved; these rows are contextual references rather than a matched-objective ranking. The table also reports the \method{} ablations analyzed below.

\subsection{Utility Degradation Across Victims}
\begin{figure}[!h]
\centering
\includegraphics[width=0.98\columnwidth]{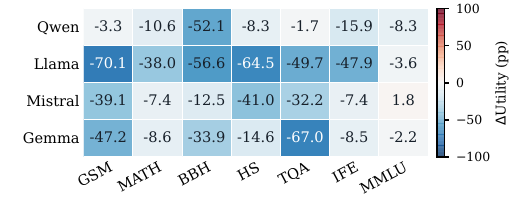}
\caption{Utility change by victim and benchmark (pp; attacked minus clean). Negative values denote degradation. HS, TQA, and IFE denote HellaSwag, TruthfulQA, and IFEval.}
\label{fig:heatmap}
\end{figure}

The aggregate utility change is \OurAvg{} percentage points (95\% CI: [\OurAvgCILow{}, \OurAvgCIHigh{}]), and \NegativeCellCount{} of \TotalCells{} victim--benchmark cells show degradation. All four victim means are negative, while Mistral on MMLU is the sole positive cell (Table~\ref{tab:main} and Fig.~\ref{fig:heatmap}). The learned prefixes therefore reduce performance on nearly all evaluated victim--benchmark cells, with the magnitude varying by victim and benchmark.

\subsection{Search and Round Selection}
In \SearchImprovementRate{} of \SearchPrompts{} victim--instruction cases, the searched winner exceeds the original best proposal after both are rescored in the same search pass; the mean gain is \SearchGain{} NLL/token. This teacher-side comparison excludes cross-pass scoring drift and measures search improvement rather than downstream utility. The search stages preceding the selected checkpoints supply \PreferencePairs{} eligible within-instruction preference pairs.

The held-out reward trajectories differ across victims (Fig.~\ref{fig:stopping}). The common rule retains $R_1$ for Qwen, Llama, and Gemma, and $R_2$ for Mistral: only the latter clears the 0.03 replacement margin.

The trajectory is non-monotonic: utility change moves from \PreFitAvg{} before fitting to \PostFitAvg{} after fitting, then reaches \OurAvg{} after refinement (Table~\ref{tab:main}). Removing the preference loop yields \NoPreferenceLoopAvg{} and permuting pseudo-references yields \ShuffledPseudoReferenceAvg{}, respectively 6.4 and 10.7~pp less negative than the complete pipeline. These matched ablations implicate both components, while request alignment is tested separately below.

\subsection{Output-Level Failure Analysis}
\begin{figure}[!h]
\centering
\includegraphics[width=0.98\columnwidth]{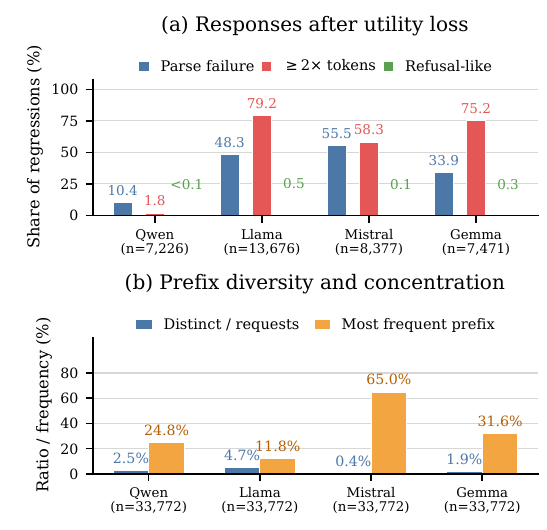}
\caption{Failure responses and prefix reuse; $n$ denotes each panel's denominator. (a) Response characteristics among requests that change from correct to incorrect. Categories overlap; refusal-like indicates a fixed lexical match. (b) Distinct-prefix ratio and frequency of the most common prefix over all requests after case and whitespace normalization.}
\label{fig:failureprofile}
\end{figure}

Among \RegressionPairs{} clean-correct/attacked-wrong victim--request pairs, \InflationShare{} contain at least twice as many output tokens and \ParseFailureShare{} contain no extracted answer under the stated parsers (Fig.~\ref{fig:failureprofile}(a)). Only \RefusalShare{} match the fixed refusal lexicon. These indicators overlap and are pooled over regression pairs, whereas the main result is an equal-cell mean; none is a semantic failure label.

The GSM8K example in Fig.~\ref{fig:motivation} illustrates task redirection: the attacked response describes solving the problem metaphorically but never performs the calculation, without a task-specific error target during training.

Prefix reuse differs markedly across victims. After case and whitespace normalization, the distinct-prefix ratio is \QwenUniqueRatio{}, \LlamaUniqueRatio{}, \MistralUniqueRatio{}, and \GemmaUniqueRatio{} for Qwen, Llama, Mistral, and Gemma, respectively (Fig.~\ref{fig:failureprofile}(b)). The most common Mistral prefix appears in \MistralTopSpanShare{} of requests, compared with \QwenTopSpanShare{} for Qwen. Prefix behavior therefore ranges from varied outputs to extensive reuse; request conditioning alone does not imply a unique prefix for every request.

The ablations and controls answer different questions. Retraining ablations hold deployment and evaluation fixed while changing the training signal; fixed-generator controls keep the learned policy fixed and change only request--prefix assignment. The former support the contribution of the preference and pseudo-reference stages under this protocol, whereas the latter leave the causal value of request matching unresolved. Together, the results support victim-side supervision and one-pass deployment without implying a uniquely tailored prefix for every request.

The fixed-generator controls provide no evidence that alignment strengthens degradation: matched-minus-shuffled is \ShuffleDelta{} pp (95\% CI [\ShuffleCILow{}, \ShuffleCIHigh{}]) and matched-minus-empty is \EmptyDelta{} pp (95\% CI [\EmptyCILow{}, \EmptyCIHigh{}]). Reusing each modal prefix increases degradation by \ModalGap{} pp (95\% CI [\ModalCILow{}, \ModalCIHigh{}]); thus, these controls do not establish an alignment advantage.
The modal gap is driven by Gemma (+6.61 pp) and Llama (+5.04 pp); Qwen and Mistral differ by less than 0.3 pp. Because this control reuses an observed generator output, it is a diagnostic of prefix concentration rather than a trained unconditional-generator baseline.

Equation~\eqref{eq:epd} uses neither benchmark answers nor a target failure string, yet the learned prefixes affect tasks with different output spaces and scoring rules. Clean continuations provide a behavioral reference, while downstream losses include wrong, missing, and instruction-violating answers on tasks excluded from attack training.

\subsection{Security Implications and Limitations}
Taken together, the experiments support white-box discovery of utility weaknesses from victim behavior, including fluent responses that abandon benign requests. Benign task performance under injected text should therefore be measured directly, not inferred from harmful compliance or refusal rates.

Response diagnostics show that refusal is rare whereas response expansion and extraction failure are common; benchmark scores should therefore be paired with direct examination of generated responses.

The generator often returns the same prefix for different inputs (Fig.~\ref{fig:failureprofile}(b)); robustness evaluation should examine both assigned pairs and frequently generated strings.

These conclusions apply to offline probability access, open-weight victims, and user-prefix insertion; closed-model transfer and indirect injection remain outside scope. One seed per victim means the intervals do not estimate retraining variability. A matched unconditional-generator branch is still needed to isolate the causal value of request input; the reported ablations test fitting and pseudo-reference permutation.

\section{Conclusion}
We introduced \method{}, which uses victim continuations to supervise one-pass prefix generation. Across four victims and seven benchmarks, the prefixes reduce benign performance without downstream labels or a prescribed failure response. Controls do not establish an alignment advantage, and prefix reuse shows that the generator can remain concentrated; the arithmetic example illustrates fluent task abandonment. These findings support white-box utility auditing alongside direct evaluation of legitimate task preservation.

\clearpage
\section*{Compliance with Ethical Standards}
This work studies a dual-use threat to benign model utility. Experiments use open-weight models and public benchmarks, with no attacks on private data, deployed services, or users. The findings can inform detection and mitigation.
Potentially harmful attack artifacts are made available only to vetted researchers under responsible disclosure.
OpenAI GPT-5.6 was used for language polishing and to assist with parts of the experimental code and visualizations. The authors reviewed, tested, and verified this assistance and remain responsible for the manuscript, code, and reported results.

\bibliographystyle{IEEEbib}
\bibliography{main}

\end{document}

%% file: results.tex
\newcommand{\TrainInstructions}{2,400}
\newcommand{\TrainSteps}{260}
\newcommand{\CleanSpanTokens}{64}
\newcommand{\ProposalTokens}{16}
\newcommand{\CandidatesPerRequest}{64}
\newcommand{\SearchTokens}{48}
\newcommand{\SearchCandidates}{512}
\newcommand{\SearchObservedMax}{515}
\newcommand{\DeploymentTokens}{16}

\newcommand{\TotalCells}{28}
\newcommand{\NegativeCellCount}{27}

\newcommand{\OurAvg}{-26.8}
\newcommand{\OurAvgCILow}{-27.4}
\newcommand{\OurAvgCIHigh}{-26.2}
\newcommand{\BootstrapReplicates}{20,000}

\newcommand{\OurQwen}{-14.3}
\newcommand{\OurLlama}{-47.2}
\newcommand{\OurMistral}{-19.7}
\newcommand{\OurGemma}{-26.0}
\newcommand{\OurQwenNegativeTasks}{7/7}
\newcommand{\OurLlamaNegativeTasks}{7/7}
\newcommand{\OurMistralNegativeTasks}{6/7}
\newcommand{\OurGemmaNegativeTasks}{7/7}

\newcommand{\PreFitAvg}{-23.6}
\newcommand{\PostFitAvg}{-5.5}
\newcommand{\NoPreferenceLoopAvg}{-20.4}
\newcommand{\ShuffledPseudoReferenceAvg}{-16.1}

\newcommand{\ShuffleDelta}{+0.21}
\newcommand{\ShuffleCILow}{-0.06}
\newcommand{\ShuffleCIHigh}{+0.48}
\newcommand{\EmptyDelta}{-0.30}
\newcommand{\EmptyCILow}{-0.65}
\newcommand{\EmptyCIHigh}{+0.06}
\newcommand{\ModalGap}{2.79}
\newcommand{\ModalCILow}{2.46}
\newcommand{\ModalCIHigh}{3.13}

\newcommand{\NegativeCells}{27/28}

\newcommand{\SearchPrompts}{9,600}
\newcommand{\PreferencePairs}{9,597}
\newcommand{\SearchImprovementRate}{96.5\%}
\newcommand{\SearchGain}{0.125}

\newcommand{\QwenUniqueRatio}{2.5\%}
\newcommand{\LlamaUniqueRatio}{4.7\%}
\newcommand{\MistralUniqueRatio}{0.4\%}
\newcommand{\GemmaUniqueRatio}{1.9\%}
\newcommand{\QwenTopSpanShare}{24.8\%}

\newcommand{\MistralTopSpanShare}{65.0\%}

\newcommand{\RegressionPairs}{36,750}
\newcommand{\ParseFailureShare}{39.5\%}
\newcommand{\InflationShare}{58.4\%}
\newcommand{\RefusalShare}{0.3\%}

\newcommand{\UATAvg}{-1.9}

\newcommand{\GCGAvg}{-7.4}

\newcommand{\BEASTAvg}{-1.4}

\newcommand{\AmpleAvg}{-4.0}

\newcommand{\AdvAvg}{-4.9}

\newcommand{\SPPAvg}{-16.2}

%% file: main.bib
@inproceedings{wallace2019universal,
  title = {Universal Adversarial Triggers for Attacking and Analyzing {NLP}},
  author = {Wallace, Eric and Feng, Shi and Kandpal, Nikhil and Gardner, Matt and Singh, Sameer},
  booktitle = {Proc. EMNLP-IJCNLP},
  pages = {2153--2162},
  doi = {10.18653/v1/D19-1221},
  year = {2019}
}

@article{parekh2021minimal,
  title = {{MINIMAL}: Mining Models for Data Free Universal Adversarial Triggers},
  author = {Parekh, Swapnil and Kumar, Yaman Singla and Singh, Somesh and Chen, Changyou and Krishnamurthy, Balaji and Shah, Rajiv Ratn},
  journal = {arXiv preprint arXiv:2109.12406},
  year = {2021}
}

@inproceedings{shin2020autoprompt,
  title = {{AutoPrompt}: Eliciting Knowledge from Language Models with Automatically Generated Prompts},
  author = {Shin, Taylor and Razeghi, Yasaman and Logan, Robert L. IV and Wallace, Eric and Singh, Sameer},
  booktitle = {Proc. EMNLP},
  pages = {4222--4235},
  doi = {10.18653/v1/2020.emnlp-main.346},
  year = {2020}
}

@inproceedings{jin2019textfooler,
  title = {Is {BERT} Really Robust? A Strong Baseline for Natural Language Attack on Text Classification and Entailment},
  author = {Jin, Di and Jin, Zhijing and Zhou, Joey Tianyi and Szolovits, Peter},
  booktitle = {Proc. AAAI Conf. Artif. Intell.},
  volume = {34},
  pages = {8018--8025},
  doi = {10.1609/aaai.v34i05.6311},
  year = {2020}
}

@inproceedings{li2020bertattack,
  title = {{BERT-ATTACK}: Adversarial Attack Against {BERT} Using {BERT}},
  author = {Li, Linyang and Ma, Ruotian and Guo, Qipeng and Xue, Xiangyang and Qiu, Xipeng},
  booktitle = {Proc. EMNLP},
  pages = {6193--6202},
  doi = {10.18653/v1/2020.emnlp-main.500},
  year = {2020}
}

@inproceedings{wen2023hardprompts,
  title = {Hard Prompts Made Easy: Gradient-Based Discrete Optimization for Prompt Tuning and Discovery},
  author = {Wen, Yuxin and Jain, Neel and Kirchenbauer, John and Goldblum, Micah and Geiping, Jonas and Goldstein, Tom},
  booktitle = {Adv. Neural Inf. Process. Syst.},
  volume = {36},
  pages = {51008--51025},
  year = {2023}
}

@article{zou2023universal,
  title = {Universal and Transferable Adversarial Attacks on Aligned Language Models},
  author = {Zou, Andy and Wang, Zifan and Carlini, Nicholas and Nasr, Milad and Kolter, J. Zico and Fredrikson, Matt},
  journal = {arXiv preprint arXiv:2307.15043},
  year = {2023}
}

@inproceedings{liao2024amplegcg,
  title = {{AmpleGCG}: Learning a Universal and Transferable Generative Model of Adversarial Suffixes for Jailbreaking Both Open and Closed {LLMs}},
  author = {Liao, Zeyi and Sun, Huan},
  booktitle = {Proc. Conf. Lang. Model. (COLM)},
  year = {2024}
}

@inproceedings{paulus2024advprompter,
  title = {{AdvPrompter}: Fast Adaptive Adversarial Prompting for {LLMs}},
  author = {Paulus, Anselm and Zharmagambetov, Arman and Guo, Chuan and Amos, Brandon and Tian, Yuandong},
  booktitle = {Proc. Int. Conf. Mach. Learn. (ICML)},
  volume = {267},
  pages = {48439--48469},
  year = {2025}
}

@inproceedings{sadasivan2024beast,
  title = {Fast Adversarial Attacks on Language Models In One {GPU} Minute},
  author = {Sadasivan, Vinu Sankar and Saha, Shoumik and Sriramanan, Gaurang and Kattakinda, Priyatham and Chegini, Atoosa and Feizi, Soheil},
  booktitle = {Proc. Int. Conf. Mach. Learn. (ICML)},
  volume = {235},
  pages = {42976--42998},
  year = {2024}
}

@inproceedings{proadvprompter2025,
  title = {{ProAdvPrompter}: A Two-Stage Journey to Effective Adversarial Prompting for {LLMs}},
  author = {Di, Hao and others},
  booktitle = {Proc. Int. Conf. Learn. Represent. (ICLR)},
  year = {2025}
}

@inproceedings{zhu2023promptbench,
  title = {{PromptRobust}: Towards Evaluating the Robustness of Large Language Models on Adversarial Prompts},
  author = {Zhu, Kaijie and others},
  booktitle = {Proc. ACM Workshop Large AI Syst. Models Privacy Saf. Anal.},
  pages = {57--68},
  doi = {10.1145/3689217.3690621},
  year = {2024}
}

@inproceedings{cao2024worstprompt,
  title = {On the Worst Prompt Performance of Large Language Models},
  author = {Cao, Bowen and Cai, Deng and Zhang, Zhisong and Zou, Yuexian and Lam, Wai},
  booktitle = {Adv. Neural Inf. Process. Syst.},
  volume = {37},
  pages = {69022--69042},
  doi = {10.52202/079017-2205},
  year = {2024}
}

@inproceedings{greshake2023indirect,
  title = {Not What You've Signed Up For: Compromising Real-World {LLM}-Integrated Applications with Indirect Prompt Injection},
  author = {Greshake, Kai and Abdelnabi, Sahar and Mishra, Shailesh and Endres, Christoph and Holz, Thorsten and Fritz, Mario},
  booktitle = {Proc. ACM Workshop Artif. Intell. Secur. (AISec)},
  pages = {79--90},
  doi = {10.1145/3605764.3623985},
  year = {2023}
}

@inproceedings{zhang2024safeguarddos,
  title = {{LLM} Safeguard is a Double-Edged Sword: Exploiting False Positives for Denial-of-Service Attacks},
  author = {Zhang, Qingzhao and Xiong, Ziyang and Mao, Z. Morley},
  booktitle = {Proc. Workshop Large AI Syst. Models Privacy Secur. Anal.},
  pages = {1--10},
  doi = {10.1145/3733800.3763264},
  year = {2025}
}

@article{li2025systempromptpoisoning,
  title = {System Prompt Poisoning: Persistent Attacks on Large Language Models Beyond User Injection},
  author = {Li, Zongze and Guo, Jiawei and Cai, Haipeng},
  journal = {arXiv preprint arXiv:2505.06493},
  year = {2025}
}

@inproceedings{hendrycks2020measuring,
  title = {Measuring Massive Multitask Language Understanding},
  author = {Hendrycks, Dan and others},
  booktitle = {Proc. Int. Conf. Learn. Represent. (ICLR)},
  year = {2021}
}

@inproceedings{suzgun2022challenging,
  title = {Challenging {BIG-Bench} Tasks and Whether Chain-of-Thought Can Solve Them},
  author = {Suzgun, Mirac and others},
  booktitle = {Findings ACL},
  pages = {13003--13051},
  doi = {10.18653/v1/2023.findings-acl.824},
  year = {2023}
}

@inproceedings{cobbe2021training,
  title = {Training Verifiers to Solve Math Word Problems},
  author = {Cobbe, Karl and others},
  booktitle = {Proc. Int. Conf. Learn. Represent. (ICLR)},
  year = {2022}
}

@inproceedings{hendrycks2021measuringmath,
  title = {Measuring Mathematical Problem Solving With the {MATH} Dataset},
  author = {Hendrycks, Dan and others},
  booktitle = {Adv. Neural Inf. Process. Syst.},
  volume = {34},
  year = {2021}
}

@inproceedings{zellers2019hellaswag,
  title = {{HellaSwag}: Can a Machine Really Finish Your Sentence?},
  author = {Zellers, Rowan and Holtzman, Ari and Bisk, Yonatan and Farhadi, Ali and Choi, Yejin},
  booktitle = {Proc. ACL},
  pages = {4791--4800},
  doi = {10.18653/v1/P19-1472},
  year = {2019}
}

@inproceedings{lin2022truthfulqa,
  title = {{TruthfulQA}: Measuring How Models Mimic Human Falsehoods},
  author = {Lin, Stephanie and Hilton, Jacob and Evans, Owain},
  booktitle = {Proc. ACL},
  pages = {3214--3252},
  doi = {10.18653/v1/2022.acl-long.229},
  year = {2022}
}

@article{zhou2023instruction,
  title = {Instruction-Following Evaluation for Large Language Models},
  author = {Zhou, Jeffrey and others},
  journal = {arXiv preprint arXiv:2311.07911},
  year = {2023}
}

@article{shao2024deepseekmath,
  title = {{DeepSeekMath}: Pushing the Limits of Mathematical Reasoning in Open Language Models},
  author = {Shao, Zhihong and others},
  journal = {arXiv preprint arXiv:2402.03300},
  year = {2024}
}

@inproceedings{azar2024ipo,
  title = {A General Theoretical Paradigm to Understand Learning from Human Preferences},
  author = {Azar, Mohammad Gheshlaghi and others},
  booktitle = {Proc. Int. Conf. Artif. Intell. Stat. (AISTATS)},
  pages = {4447--4455},
  volume = {238},
  year = {2024}
}

@misc{conover2023dolly,
  title = {Free Dolly: Introducing the World's First Truly Open Instruction-Tuned {LLM}},
  author = {Conover, Mike and others},
  howpublished = {Databricks Blog},
  year = {2023}
}

@article{qwen25technical,
  title = {{Qwen2.5} Technical Report},
  author = {Yang, An and others},
  journal = {arXiv preprint arXiv:2412.15115},
  year = {2024}
}

@article{llama3herd,
  title = {The {Llama 3} Herd of Models},
  author = {Grattafiori, Aaron and others},
  journal = {arXiv preprint arXiv:2407.21783},
  year = {2024}
}

@article{mistral7b,
  title = {{Mistral 7B}},
  author = {Jiang, Albert Q. and others},
  journal = {arXiv preprint arXiv:2310.06825},
  year = {2023}
}

@article{gemma2report,
  title = {{Gemma 2}: Improving Open Language Models at a Practical Size},
  author = {{Gemma Team} and Riviere, Morgane and others},
  journal = {arXiv preprint arXiv:2408.00118},
  year = {2024}
}

@misc{dolphin30llama32,
  title = {{Dolphin 3.0 Llama 3.2 3B}},
  author = {Hartford, Eric and Gitter, Ben and BlouseJury and {{Cognitive Computations}}},
  howpublished = {Hugging Face model card, dphn/Dolphin3.0-Llama3.2-3B},
  year = {2025}
}
